%% file: wsdm2016.tex
\documentclass{sig-alternate-2013}
\usepackage[ruled]{algorithm2e}
\usepackage{subfigure}
\usepackage{multirow}
\usepackage{color, verbatim,caption}
\usepackage{hyperref}
\newtheorem{definition}{Definition}

\newfont{\mycrnotice}{ptmr8t at 7pt}
\newfont{\myconfname}{ptmri8t at 7pt}
\let\crnotice\mycrnotice%
\let\confname\myconfname%

\permission{Permission to make digital or hard copies of all or part of this work for personal or classroom use is granted without fee provided that copies are not made or distributed for profit or commercial advantage and that copies bear this notice and the full citation on the first page. Copyrights for components of this work owned by others than ACM must be honored. Abstracting with credit is permitted. To copy otherwise, or republish, to post on servers or to redistribute to lists, requires prior specific permission and/or a fee. Request permissions from Permissions@acm.org.}
\conferenceinfo{xxxxx,}{xxxxxxx.} 

\begin{document}

\title{Identity-sensitive Word Embedding through Heterogeneous Networks}

\numberofauthors{3}
\author{	
\alignauthor
Jian Tang\\
       \affaddr{Microsoft Research}\\
       \email{ jiatang@microsoft.com}
\alignauthor
Meng Qu\\
       \affaddr{Peking University}\\
       \email{mnqu@pku.edu.cn}
\alignauthor
Qiaozhu Mei\\
       \affaddr{University of Michigan}\\
       \email{qmei@umich.edu}
}

\maketitle

\begin{abstract}
	
Most existing word embedding approaches do not distinguish the same words in different contexts, therefore ignoring their contextual meanings. As a result, the learned embeddings of these words are usually a mixture of multiple meanings. In this paper, we acknowledge multiple identities of the same word in different contexts and learn the \textbf{identity-sensitive} word embeddings. Based on an identity-labeled text corpora, a heterogeneous network of words and word identities is constructed to model different-levels of word co-occurrences. The heterogeneous network is further embedded into a low-dimensional space through a principled network embedding approach, through which we are able to obtain the embeddings of words and the embeddings of word identities. We study three different types of word identities including topics, sentiments and categories. Experimental results on real-world data sets show that the identity-sensitive word embeddings learned by our approach indeed capture different meanings of words and outperforms competitive methods on tasks including text classification and word similarity computation.

\end{abstract}
\vskip 5pt


\input{intro}
\input{related}
\input{definition}
\input{model}
\input{experiment}

\input{conclusion}

\bibliographystyle{abbrv}

\small{
\bibliography{sigproc}
}
\end{document}

%% file: intro.tex
\section{Introduction}
\label{sec::intro}
Learning a meaningful representation of words is critical in many text mining tasks. 
Traditional vector space models represent a word as a ``one-hot'' vector in a high-dimensional space, which suffers from serious problems of data sparseness. Recent approaches represent words as real vectors in a low-dimensional space (i.e., word embedding), which much better capture the syntactic and semantic relationships between the words. Word embeddings have been proved to be very effective in many tasks such as word analogy~\cite{mikolov2013distributed}, named entity recognition~\cite{collobert2011natural}, POS tagging~\cite{collobert2011natural}, and text classification~\cite{maas2011learning,tang2015pte,kim2014convolutional}.

The basic intuition of learning word embeddings comes from the distributional hypothesis that ``you shall know a word by the company it keeps" (Firth, J.R. 1957:11)~\cite{Firth1957}. Such an intuition has motivated many models, including the first word embedding approach based on neural networks~\cite{bengio2003neural}, SENNA~\cite{collobert2011natural}, Glove~\cite{pennington2014glove}, Skipgram~\cite{mikolov2013distributed}, etc. Among these models, the Skipgram model has been widely adopted due to its simplicity, effectiveness and efficiency. The basic idea of the Skipgram is to use the embedding of the target word to predict the embeddings of the surrounding context words in local context windows. 

Although these existing approaches have been proved to be already quite effective in some tasks, a common problem of them is that they do not distinguish the meanings of the same words in different contexts. In other words, every word is represented with a single embedding vector, even if they have multiple, clearly different meanings (e.g., ``java,'' ``bank''). Some words have multiple senses and each corresponds to a different meaning (e.g., ``tear'' as a noun or as a verb). Some words have one sense but their meanings vary subtly in different contexts (e.g., ``network'' in the data mining community or in the computer networks community). Representations of these words learned by existing word embedding approaches are a mixture of their multiple meanings, thus compromising the accuracy of semantic relations. To address this problem, the contextual meanings of words must be taken into consideration.

\begin{figure*}[htdb!]
	\centering
	\includegraphics[width=0.75\textwidth]{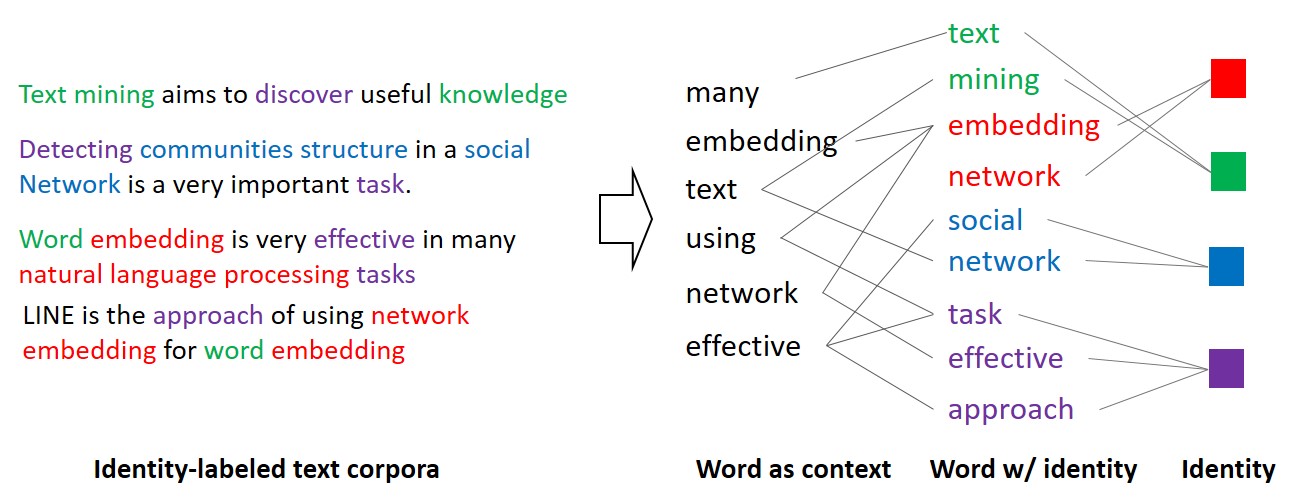}
	\caption{Illustration of converting a identity-labeled text corpora to a heterogeneous network. Different colors represent different word identities. The heterogeneous network encodes the local context-level word co-occurrences through the word-context bipartite network. When constructing the word-context network, the identities of words are ignored if the words are treated as contexts. The heterogeneous network also encodes identity-level word co-occurrences through the word-identity bipartite network.}
	\label{fig::text2netwok}
\end{figure*}

In this paper, we acknowledge that same word may take on different \textbf{identities} when they are used in different contexts, and every such identify carries its own meaning and should be represented with its own embedding. In other words, we learn the \textbf{identity-sensitive} word embeddings. Our intuition is inspired by the theories of social identity and self categorization  in social psychology and economics. These theories suggest that every person can sense multiple identities of herself, which are usually perceived from her membership in different social groups, and some identity may dominate other identities under particular situations \cite{tajfel1974social,turner1987rediscovering}. For example, the first author may perceive himself as both a Microsoft employee, a data mining researcher, and a University of Michigan alumni, and the third identity becomes dominant in a conversation about Ohio States. Following a simple analogy, we define the identity of a word as its membership belonging to a group of words, which may be derived from a taxonomy, a topic modeling process~\cite{blei2003latent}, or any other meaningful partitions of words. Besides topic identity, there may be many other types of word identities such as categories, sentiments, POS tags and ideologies. The meanings of a word with different identities are usually different. 
Therefore, a sound word embedding should allow each word to have a different embedding for every identity it belongs to, i.e., to have identity-sensitive embeddings.

To learn the identity-sensitive word embeddings, we propose an approach based on heterogeneous network embedding. Given a text collection, we first recognize the identities of \textbf{every word token} in that corpus, therefore the same word in different contexts may be labeled with different identities. This identity recognition process may take any reasonable existing word labeling approaches and vary for different types of identities. For example, the topic identities of word tokens can be recognized through the Gibbs sampling algorithm for the latent Dirichlet allocation~\cite{blei2003latent}; the part-of-speech of words can be obtained though a sequence labeling method such as the conditional random fields~\cite{lafferty2001conditional}; the sentiment polarity of word tokens can be obtained through a sentiment dictionary and a word-sense disambiguation process. 
Based on the identity-labeled corpus, we build a heterogeneous network consisting of words and word identities, which encodes different levels of essential word co-occurrence information for the embedding process. The heterogeneous network contains a word-context network that encodes the local context level word co-occurrences, which are mainly utilized by existing word embedding approaches such as Skipgram; it also contains a word-identity bipartite network that encodes identity-level word co-occurrences. The intuition is that words with the same identity are likely to take similar meanings. The network constructing process is illustrated in Fig.~\ref{fig::text2netwok}. We extend our previous work on homogeneous network embedding~\cite{tang2015line} to learn the embedding of the heterogeneous network, which is very efficient and scalable.

We present comprehensive experiments on three different types of word identities including topics, sentiments and categories. The effectiveness of identity-sensitive word embeddings is carefully evaluated using the tasks of text classification and word similarity computation. Experimental results show that the identity-sensitive word embeddings are indeed able to capture different meanings of the same words, outperforming competitive approaches including some state-of-the-art word embedding approaches and other methods that consider the contextual meanings of words.

To summarize, we make the follow contributions:

\begin{itemize}
	\item We study identity-sensitive word embeddings which is inspired by social identity and self categorization theories. We acknowledge that the same words may take on different identities  and present different meanings in different contexts.
	\item We use a heterogeneous network to encode different levels of co-occurrence information between words and word identities. A principled heterogeneous network embedding approach is proposed to learn the embeddings of words and word identities.
	\item We study three different types of word identities including topics, sentiments and categories. Experimental results show that our proposed approach outperforms existing methods on real tasks and real data sets.
\end{itemize}

The rest of this paper is organized as follows. Section~\ref{sec::related} discusses the related work. Section~\ref{sec::definition} formally defines the problem of identity-sensitive word embedding via heterogeneous networks. Section~\ref{sec::model} describes our approach of heterogeneous network embedding. Section~\ref{sec::experiment} presents the experimental results, and we conclude in Section~\ref{sec::conclusion}.

%% file: related.tex
\section{Related Work}
\label{sec::related}

Word embedding has been attracting increasing attention recently~\cite{bengio2003neural, mnih2007three, mnih2009scalable,collobert2011natural, huang2012improving,mnih2013learning,mikolov2013distributed,pennington2014glove,tang2015line}. Each word is represented with a low-dimensional vector, which can be used as features for a variety of tasks. The first word embedding approach is proposed by Bengio et al.~\cite{bengio2003neural} for language modeling purpose, in which a neural network is built for predicting the next word given the previous words in a sentence. Mnih et al.~\cite{mnih2007three} further simplified the neural language model with a log-bilinear model. However, both of the two previous approaches suffer from serious problem of high computational complexity due to the large model output space (the vocabulary). Mnih et al.~\cite{mnih2009scalable} addressed the problem through the hierarchical softmax techniques. Mikolov et al.~\cite{mikolov2013distributed} proposed a very simple model Skipgram by just using the embedding of the target word to predict the embeddings of the surrounding context words in local context windows, which is essentially using the local context-level word co-occurrences. Compared to the Skipgram, our previous work~\cite{tang2015line} showed that better results can be obtained through constructing the word co-occurrence network first and then embedding the co-occurrence network.

Most existing embedding approaches represent each word with a single embedding vector, ignoring the variation of word meanings in different contexts. A more sound solution in practice is to represent each word with multiple embeddings, each of which captures a different meaning of the word. There are indeed some recent work that learns multiple embeddings for each word~\cite{huang2012improving,chen2014unified,neelakantan2015efficient,liu2015topical}. Huang et al.~\cite{huang2012improving} used a post-processing approach to cluster the context vectors of each word, and then each context cluster is treated as a embedding of the word.
Neelakatan et al.~\cite{neelakantan2015efficient} proposed a multi-sense Skipgram (MSSG) model to estimate multiple embedding vectors for each word during the learning process. Each sense of the word is associated with a context cluster. During the learning process, the average context vectors is compared with the context cluster centers to determine the sense of the target word. In the experiments, we compare with the MSSG model. Chen et al.~\cite{chen2014unified} used external knowledge WordNet to estimate an embedding for each sense of a word. Compared with their work, our approach does not require external knowledge. The most similar work to us is~\cite{liu2015topical}, in which topic identity is studied to differentiate the meanings of words in different contexts. However, in their work only topic identity is studied while we studied multiple types of word identities besides topic identity. Besides, all previous approaches learned the embeddings from free text while our approach learned the embeddings through embedding heterogeneous networks, which are able to capture different levels of word co-occurrences. 

%% file: definition.tex
\section{Problem Definition}
\label{sec::definition}
The meanings of words are usually ambiguous, depending on the contexts. Most existing word embedding approaches ignore the variations of word meanings in different contexts. As a result, each word is represented with a single embedding vector. In this paper, we take the initiative to study context-aware word embedding. We acknowledge that the same word may have different \textbf{identities} in different contexts. We define the identity of word as follows:

\begin{definition}
	\label{def::identity}
	\textbf{(Word Identity)}
	\textsl{The \textbf{identity} of a word is its membership belonging to a group of words. For example, the word ``cat'' has the identity of ``animal'' as it belongs to the ``animal''; the word ``flower'' has the identity of ``plant'' as it belongs to the ``plant''.}
\end{definition}

There are many types of word identities, e.g., topics, sentiments, ideologies, categories and POS tags. For each type of identity, each word may take on multiple identities. Take the topic identity as an example, each word may belong to multiple topics; another example is the sentiment identity, some words can have both the ``positive'' and ``negative'' sentiment orientation. When a word takes on different identities, the meanings of the word are usually different. Therefore, each word with a different identity should be represented with a different embedding, capturing a different meaning of the word, i.e., identity-sensitive word embedding. Specifically, we define identity-sensitive word embedding as follows:

\begin{definition}
	\label{def::indentity-embedding}
	\textbf{(Identity-sensitive Word Embedding)}
	\textsl{\textbf{Identity-sensitive} word embedding means that each word $w$ has a different embedding $\vec{w}_i$ for each identity $i\in　I $, where $I$ is the set of word identities.}
\end{definition}

Note that although we assume that each word has a different embedding for each identity. In practice, each word usually has a few identities. For example, each word usually belongs to a few topics; many words only take on either positive or negative sentiment orientation. Therefore, compared to the word embeddings without considering the word identities, the number of parameters for identity-sensitive word embeddings will not increase significantly. 

To learn the identity-sensitive word embeddings, the identities of the tokens in the corpus must be recognized first. The recognizing process depends on the types of identities. For example, the topic identities can be recognized through the Gibbs sampling of the latent Dirichlet allocation (LDA) model; the sentiment identities can be determined by the sentiment identities of the sentences. Once the identities of the tokens in the corpora are recognized, we can leverage different levels of word co-occurrences, which are the essential information for word embedding. One intuition is based on the distributional hypothesis, which assumes that words co-occur with similar words are likely to have similar meanings. Therefore, the local context-level word co-occurrences can be leveraged, which are also used by the Skipgram model. To encode the local context-level word co-occurrences, we introduce the word-context bipartite network:


\begin{definition}
	\label{def::wc}
	\textbf{(Word-context Network $G_{wc}$)}
	\textsl{Word-context network, denoted as $G_{wc}=(V_w\cup V_{c}, E_{wc})$, is a bipartite network. $V_{w}$ is the set of words with identities, $V_c$ is the vocabulary of context words without considering the word identities.  $E_{wc}$ is the set of edges between words with identities and context words. The weight of the edge between word $w$ and context word $c$ is the number of times they co-occur in all the local context windows. }
\end{definition}

Note that when constructing the word-context network, the identities of the words are ignored when the words are treated as contexts. The reason is that by doing this, the number of unique context words can be reduced, decreasing the data sparsity. Besides the local context-level word co-occurrences, another intuition is that words with the same identity are likely to take on similar meanings. For example, words assigned to the same topic are likely to take similar meanings. Therefore, the identity-level word co-occurrences can be leveraged. To encode the identity-level word co-occurrences, we introduce the word-identity network, defined as below:

\begin{definition}
	\label{def::wi}
	\textbf{(Word-Identity Network $G_{wi}$)}
	\textsl{Word-identity network, denoted as $G_{wi}=(V_{w}\cup V_I, E_{wi})$, is also a bipartite network, where $V_I$ is the set of identities. $E_{wi}$ is the set of edges between words with identities and word identities. The weight between word $w$ and identity $i$ is the number of times word $w$ taking on identity $i$ in the entire corpus. }
\end{definition}

The word-context network and word-identity network encodes different levels of word co-occurrences. To leverage both levels of word co-occurrences, we further combine the two networks into a heterogeneous network. Therefore, to learn the word embeddings, we can embed the heterogeneous network constructed from the identity-labeled corpus. Formally, we define our problem as follows:

\begin{definition}
	\label{def::problem definition}
	\textbf{(Problem Definition)}
	\textsl{Give an identity-labeled text corpora, the problem of \textbf{identity-sensitive word embedding} aims to learn multiple low-dimensional embedding vectors for each word through embedding a heterogeneous network constructed from the identity-labeled text copora. }
\end{definition}

In the next section, we introduce our approach of heterogeneous network embedding for identity-sensitive word embedding. 

%% file: model.tex
\section{Identity-sensitive Word Embedding}
\label{sec::model}
In this section, we introduce our approach of embedding heterogeneous networks, through which we can obtain both the identity-sensitive word embeddings and the embeddings of word identities. We first review our previous model LINE on large-scale information network embedding~\cite{tang2015line}. 

The LINE model is applicable for embedding arbitrary types of networks including directed, undirected and weighted. It is mainly designed for homogeneous information network embedding, in which there is only one type of nodes in the network. To learn a low-dimensional representation of the vertices in the network, the intuition is to preserve the relationship or proximity between the vertices in the low-dimensional spaces. The essential idea of LINE is to preserve the first-order or second-order proximities between the vertices, which assumes that vertices with links or vertices with similar neighbors are similar to each other. Based on the distributional hypothesis in linguistics, words with similar company or neighbors are likely to be similar to each other. This corresponds to the assumption of the second-order proximity in the word-context network. Therefore, we only utilize the second-order proximity here. 

Next, we introduce how to utilize the LINE model for bipartite network embedding. Given a bipartite network $G=(V_1\cup V_2, E_{12})$, where $V_1$ and $V_2$ are two disjoint sets of vertices, and $E_{12}$ are the set of edges between the two sets of vertices. For each edge $e=(i,j), i\in V_1, j\in V_2$, the probability of vertex $v_j$ conditioned on vertex $v_i$ can be defined as follows:

\begin{equation}
\label{eqn::cond-prob}
p(v_j|v_i)= \frac{\exp(\vec{v}_j^T\cdot \vec{v}_i)}{\sum_{k\in V_2}\exp(\vec{v}_k^T\cdot \vec{v}_i)},
\end{equation}
where $\vec{v}_i$ is the low-dimensional representation of vertex $v_i$. 

Given a vertex $v_i\in V_1$, we can see that Eqn.~\eqref{eqn::cond-prob} actually defines a conditional distribution over the vertices in the set $V_2$. In practice, the second-order proximity between a pair of vertex $(v_i,v_{i'})$ can actually be determined by the corresponding empirical conditional distribution $\hat{p}(\cdot|v_i)$ and $\hat{p}(\cdot|v_{i'})$. Therefore, to preserve the second-order proximity, we can make the conditional distribution $p(\cdot|v_i)$ be close to the empirical distribution $\hat{p}(\cdot|v_i)$. The final objective function can be formulated as follows:

\begin{equation}
	\label{eqn::obj}
	O=\sum_{i\in V_1} \lambda_i d(\hat{p}(\cdot|v_i), p(\cdot|v_i) ),
\end{equation}
where $d(\cdot,\cdot)$ is the distance function between two probability distributions, and here we adopt the widely used distance of KL-divergence. The empirical probability can be defined as $\hat{p}(v_j|v_i)=\frac{w_{ij}}{d_i}$, where $d_i$ is the degree of vertex $v_i$. $\lambda_i$ is the prestige of vertex $v_i$ in the network, and we use a very simple approach by setting $\lambda_i=\sum_{k}w_{ik}$, i.e., the degree of vertex $v_i$. Removing some constants, the final objective function~\eqref{eqn::obj} can be simplified as below:

\begin{equation}
\label{eqn::obj-final}
O=-\sum_{(i,j)\in E_{12}} w_{ij} \log p(v_j|v_i).
\end{equation} 

As mentioned above, the identity-sensitive word embeddings can be obtained through embedding the heterogeneous network constructed from the identity-labeled text corpora. The heterogeneous network includes two bipartite networks: word-context network and word-identity network. Therefore, a intuitive approach is to minimize the following objective function:

\begin{equation}
	\label{eqn::obj-hn}
	\small
	O_{HN}=O_{wc}+O_{wi},
\end{equation}
where $O_{wc}$ is the objective function for word-context network and defined as:
\begin{equation}
\label{eqn::obj-wc}
\small
O_{wc}=-\sum_{(j,k)\in E_{wc}} w_{jk} \log p(w_j|c_k),
\end{equation}
and $O_{wi}$ is the objective function for word-identity network and defined as:
\begin{equation}
\label{eqn::obj-wc}
\small
O_{wi}=-\sum_{(j,k)\in E_{wi}} w_{jk} \log p(w_j|i_k).
\end{equation}

Note that for the objective function~\eqref{eqn::obj-hn}, a better approach may be weighting the two networks differently. However, we found that equally weighting the two networks consistently perform very well in different data sets, and this is able to save users' effort in choosing a optimum parameter for controlling the importance of the two networks. 

\textbf{Optimization.}
Next, we describe the solution to optimize the objective function~\eqref{eqn::obj-hn}. The optimization can be approached through the techniques of negative sampling~\cite{mikolov2013distributed} and edge sampling~\cite{tang2015line}. For each vertex $v_i \in V_1$, the conditional probability~\eqref{eqn::cond-prob} requires calculating the similarities between vertex $v_i$ and all the vertices in the set $V_2$, which is very computationally expensive. The negative sampling technique approaches this through drawing some negative edges for each positive edge. Specifically, it aims to minimize the following objective function:

\begin{eqnarray}
\label{eqn::ns}
\scriptsize
\begin{aligned}
	O_{ns}=&-\sum_{(j,k)\in E_{wc}} w_{jk} (\log \sigma(\vec{w}_j^T \cdot \vec{c}_k) + K\sum_{n\sim P_n(w)} \log \sigma(-\vec{w}_n^T\cdot \vec{c}_k))\\
	&-\sum_{(j,k)\in E_{wi}} w_{jk} (\log \sigma(\vec{w}_j^T\cdot \vec{i}_k ) + K\sum_{n\sim P_n(w)} \log \sigma(-\vec{w}_n^T\cdot \vec{i}_k)),  
\end{aligned}
\end{eqnarray}
where $K$ is the number of negative samples for each positive edge. $\sigma(\cdot)=1/(1+\exp(-x))$ is the sigmoid function. $P_n(w)$ is the noisy distribution of words, which is used to generating negative edges. $P_n(w)$is usually set as $P_n(w)\propto t(w)^{0.75}$ according to~\cite{mikolov2013distributed,tang2015line}, where $t(w)$ is the frequency of the word $w$. 

The objective function~\eqref{eqn::ns} can be optimized through stochastic gradient descent. When an edge $(i,j)$ is sampled, the weight of the edge $w_{ij}$ will be multiplied into the gradients. This is problematic when the values of the weights diverge, i.e., some weights are very large while some are very small. In this case, it is very hard to choose an appropriate learning rate when different edges are sampled for model updating. The edge sampling~\cite{tang2015line} approach effectively addresses this through sampling the edges according to their probabilities and treating the sampled edges as binary edges. 

When using edge sampling for optimizing objective function~\eqref{eqn::ns}, one concern is that the weights of the edges in the two networks $G_{wc}$ and $G_{wi}$ may not be comparable. This can be resolved through alternatively sampling the edges from the two networks. The overall learning process can be summarized into Alg.~\ref{algo::training}. 

\begin{algorithm}[!htdb]
	\scriptsize
	\KwData{$G_{wc},G_{wi}$, number of samples $T$, number of negative samples $K$.}
	\KwResult{identity-sensitive word embeddings, embeddings of word identities.}
	\While{ iter $\leq$ $T$}{
		\begin{itemize}
			\item	draw a positive edge from $E_{wc}$ and $K$ negative edges according to the noisy distribution $P_n(w)$, and update \\ the identity-sensitive word embeddings and context embeddings\;
			\item	draw a positive edge from $E_{wi}$ and $K$ negative edges according to the noisy distribution $P_n(w)$, and update \\ the identity-sensitive word embeddings and the \\ embeddings of word identities\;
		\end{itemize}
	}
	\caption{Training process of identity-sensitive word embedding.}
	\label{algo::training}
\end{algorithm}

%% file: experiment.tex
\section{Experiments}
\label{sec::experiment}
In this section, we move forward to evaluate the effectiveness of our proposed approach on real-world data sets. Three types of word identities are studied including \emph{topics}, \emph{sentiments} and \emph{categories}. The identity-sensitive word embeddings are evaluated through the tasks of text classification and contextual word similarity measuring. We first introduce our experiment setup.

\subsection{Experiment Setup}

\subsubsection{Data Sets}
\begin{table*}[!htdb]
	\label{tab::datasets}
	\centering
	\caption{Statistics of the data sets.}
	\scalebox{1}{
		\begin{tabular}{c|c|c|c|c|c|c|c} \hline
			& \textsc{20NG}& \textsc{DBLP} & \textsc{WikiSample}& \textsc{WikiFull}& \textsc{Twitter} & \textsc{MR} &\textsc{TreeBank} \\ \hline
			\# Train& 11,314 & 25,026  & 42,000&  3,405,189    & 800,000        & 7,108     &7,792      \\ \hline
			\# Test & 7,532  & 12,520   &  21,000&   -   & 400,000        & 3,554     & 1,821      \\ \hline
			Doc.Length&305.77  & 77.04  &672.56  &   224.20   &    14.36     &    22.02    & 20.28  \\ \hline
			\#Category&20 & 8 &  7 &-    &      2   &    2     & 2 \\ \hline
			Identity & topic, category& topic, category & topic, category &topic &sentiment & sentiment & sentiment \\ \hline
		\end{tabular}
	}
	\label{tab::datasets}
\end{table*}

\begin{itemize}
	\item \textsc{20NG}, the widely used data set 20newsgroup for text mining\footnote{Available at~\url{http://qwone.com/~jason/20Newsgroups/}}. On this data set, two types of word identities \emph{topic} and \emph{category} are recognized.
	\item \textsc{DBLP}, the abstracts of the papers in computer science bibliography\footnote{Available at ~\url{https://aminer.org/billboard/citation}}. In this data set, 8 diverse research fields are used including ``data mining'', ``database'', ``artificial intelligence'', ``computational linguistics'', ``hardward'', ``system'', ``programming language'' and ``theory''. The \emph{topic} and \emph{category} identities are also recognized on this data set. 
	\item \textsc{WikiSample}, a collection of sampled English Wikipedia articles and used in~\cite{tang2015pte}.  Seven diverse categories from the DBpedia ontology are selected including ``Arts'', ``History'', ``Human'', ``Mathematics'', ``Nature'', ``Technology'' and ``Sports''. For each category, 9,000 articles are randomly selected into the training sets. On this data set, the \emph{topic} and \emph{category} identities are recognized. 
	\item \textsc{WikiFull}, the entire English Wikipedia articles\footnote{Available at~\url{https://en.wikipedia.org/wiki/Wikipedia:Database_download}}. On this data set, the \emph{topic} identity is recognized. This data set is used for the task of contextual word similarity measuring.
	\item \textsc{Twitter}, a large collection of tweets used for sentiment analysis~\footnote{\url{Available at http://thinknook.com/ twitter-sentiment-analysis-training-corpus-dataset-\\2012-09-22/}}. 120,000 tweets are sampled and split into training and test data sets. On this data set, the \emph{sentiment} identity is recognized. 
	\item \textsc{MR}, a movie review data set from~\cite{pang2005seeing}.
	\item \textsc{Treebank}, the Stanford sentiment Treebank dataset\footnote{Available \url{http://nlp.stanford.edu/sentiment/index.html}}. The data set for binary sentiment classification is used. 
\end{itemize}

In all the above data sets, the stop words are removed. We summarize the statistics of these data sets in Table~\ref{tab::datasets}.

\subsubsection{Recognizing the Identities of Words}
To learn the identity-sensitive word embeddings, the identities of the word tokens in different contexts must be recognized in advance. In this part, we introduce the identity recognizing approaches for different types of identities. 

\noindent \textbf{Topic.} The topic identities of the tokens in the corpus are estimated through the Gibbs sampling algorithm~\cite{griffiths2004finding} of the latent Dirichlet allocation. 

\noindent \textbf{Sentiment.} The sentiment identities of the tokens in each \emph{training} document are determined through the identity of the document. Specifically, if a document has positive sentiment, then all the tokens in the document are assigned to the positive identity. However, not all the words have the sentiment orientations. Therefore, some feature selection methods are first used to select a set of words that are likely to have the sentiment orientation. When recognizing the identities of the tokens in a document, only the words selected by the feature selection methods are assigned to the identity of the document. 
 
For feature selection, a very simple approach is used. For each word $w$, we calculate its probability appearing in positive sentiment $p(w|pos)$ and negative sentiment $p(w|neg)$. Then words with $\frac{p(w|pos)}{p(w|neg)}$ or $\frac{p(w|neg)}{p(w|pos)}$ larger than a threshold are selected. We empirically choose the threshold as 10. 

The sentiment identities of the tokens in each of the \emph{test} document can be determined through the word embeddings learned on the \emph{training} documents. Note that each word has multiple word embeddings associated with different identities and meanwhile has a context embedding when the word is treated as a context. Given a test document, we recognize the identity of token $w$ in the document as follows:
\begin{equation}
\label{eqn::id-est}
k^*=\text{argmax}_{k} \vec{w}_k^T \vec{c}_{\bar{w}},
\end{equation}
where $\vec{w}_k$ is the embedding of word $w$ with identity $k$, $\vec{c}_{\bar{w}}$ is the average context embedding of all the other words except $w$ in the document .  

\noindent \textbf{Category.} For the category identity, the identities of the tokens in each \emph{training} document is assigned as the identity of the document. The identities of the tokens in the \emph{test} documents can be recognized in the same way as the sentiment identity according to Eqn.~\eqref{eqn::id-est}.

\subsubsection{Compared Algorithms}

\begin{itemize}
	\item BOW: the classical ``bag-of-words'' representation. Each document is represented with a $|V|$-dimensional vector, with each dimension corresponding to a word. The weight of each word/dimension is calculated according to the TFIDF weighting~\cite{salton1988term}.
	\item Skipgram: the widely used word embedding algorithm Skipgram~\cite{mikolov2013distributed}.  In the Skipgram model, words in different contexts are treated as the same words, and hence each word is only represented with a single embedding vector. 
	\item LINE: our previous work on large-scale information network embedding~\cite{tang2015line}. The LINE model can be applied for learning word embeddings through building the word co-occurrence network first and then embedding the word co-occurrence network. Words in different contexts are also not distinguished. 
	\item LDA: the latent Dirichlet allocation model~\cite{blei2003latent}. 
	\item MSSG: the multi-sense Skipgram model proposed in~\cite{neelakantan2015efficient}. The MSSG model dynamically learns multiple embeddings for each word during the training process, each of which captures a different sense of the word. In the MSSG model, we set the number of senses as 3 by default according to~\cite{neelakantan2015efficient}.
	\item TWE: the topical-word embedding model proposed in~\cite{liu2015topical}, in which topic is used as the identities of the words. The TWE model can also be used when other types of identities are recognized. We use the second variant of the TWE~\cite{liu2015topical} as it allows each word to have different embeddings for different identities. 
	\item ISE. Our solution for identity-sensitive word embedding. A heterogeneous network is constructed from the identity-labeled text corpora according to Section~\ref{sec::definition} and then the heterogeneous network is embedded according to Section~\ref{sec::model}.
\end{itemize}

\subsubsection{Evaluation}
We evaluate different word embeddings through the tasks of text classification and contextual word similarity measuring.

\noindent \textbf{Text classification.} All the models are trained on the training data sets to learn the word embeddings and evaluated on the test data sets.
For the embeddings of the documents on both training and test data sets, a very simple approach is used by averaging over the word vectors of the documents. Note that our goal is not to obtain the optimum document embeddings but to compare the word embeddings learned by different approaches. More advanced document embedding techniques can refer to recursive neural networks~\cite{socher2013recursive} or convolutional neural networks~\cite{kim2014convolutional}. Once the embeddings of the documents in both training and test data sets are obtained, the one-vs-rest logistic regression model in the LibLinear\footnote{http://www.csie.ntu.edu.tw/~cjlin/liblinear/} package is used for classification. We evaluate the classification performance through the metrics Micro-F1 and Macro-F1.

\noindent \textbf{Contextual Word Similarity.}  We also evaluate the quality of different word embeddings through the task of contextual word similarity measuring, which is introduced in~\cite{huang2012improving}. Given two sentences, the task aims to measure the similarity between a pair of words in the two sentences. For evaluation, we use the data set in~\cite{huang2012improving}, which includes 2,003 word pairs and their sentential contexts collected from Wikipedia. The contextual similarity scores of these word pairs are collected from Amazon Mechanical Turk. We use the \textsc{WikiFull} data set, in which the \emph{topic} identities of the tokens are recognized through the Gibbs sampling algorithm. By training on the \textsc{WikiFull} data sets, the identity-sensitive word embeddings can be obtained. Given a pair of words $(w,w')$ with their sentential contexts, we first recognize the identities of the two words in the sentential contexts according to the Eqn.~\eqref{eqn::id-est}, through which we are able to obtain the corresponding embeddings of the two words. Then the similarity between the two words is calculated as the cosine similarity of their word embeddings. Once the similarities between all pairs of words are calculated, the Spearman correlation is calculated between the similarities calculated based on the embeddings and the similarity scores given by humans.

\subsubsection{Parameter Settings}
The dimension of word embeddings for different models is set as 100 by default without explicitly specified. The window size is set as 5 for Skipgram, MSSG, TWE and when constructing the word-context network for both LINE and ISE. The learning rate of LINE and ISE is set as $\rho_t=\rho_0(1-t/T)$, where $T$ is the total number of edge samples and $\rho_0=0.025$ is the initial learning rate. The number of negative samples in all the embedding models is set as 5. The number of samples $T$ can be safely set to be very large, e.g., 500 million. 

\subsection{Quantitative Results}

	\begin{table*} [!htdb]
		\caption{Results of text classification on \textsc{20NG}, \textsc{DBLP}, and \textsc{WikiSample} data sets. Two types of identities \emph{topics} and \emph{categories} are considered. For the topic identity, the numbers of topics are  set as 80, 80, 100 on the three data sets respectively.}
		\label{tab::tc-1}
		\begin{center}
			\scalebox{1.0}{
				\begin{tabular}{c|c|c|c|c|c|c}\hline
					& \multicolumn{2}{c|}{\textsc{20NG}} & \multicolumn{2}{|c|}{\textsc{DBLP}}& \multicolumn{2}{|c}{\textsc{WikiSample}}\\\hline
					Algorithm&  Micro-F1& Macro-F1 &  Micro-F1& Macro-F1 &  Micro-F1& Macro-F1 \\ \hline \hline
					BOW& 80.88     &   79.30   &  82.37    &  82.50    &  79.95     &  80.03     \\ \hline
					LDA&70.17 &68.10 &79.00 &79.19 & 74.61 &74.61  \\ \hline 				
					Skipgram&76.82 &75.75 &80.72 &80.89 &77.92 &77.98 \\ \hline 
					LINE&78.76 &77.75 &81.41 &81.56 & 78.79 &78.85  \\ \hline 
					MSSG&75.66 &74.59 &80.06 &80.22 &77.61 &77.69  \\ \hline 
					TWE(topic)& 79.95 & 78.94&80.83&81.05 &78.17&78.18\\ \hline
					ISE(topic)&81.29 &82.00 &82.20 &82.44 &79.29 &79.30 \\  \hline 	
					TWE(category)&71.63 & 71.67&79.75&79.71 &76.47&76.47 \\ \hline
					ISE(category)&\textbf{84.64} &\textbf{83.89} &\textbf{83.65} &\textbf{83.71} &\textbf{80.46} &\textbf{80.46} \\  \hline 																										
				\end{tabular}
			}
		\end{center}
	\end{table*}

	\begin{table*} [!htdb]
		\caption{Results of text classification on the \textsc{Twitter}, \textsc{MR}, and \textsc{Treebank} data sets. The \emph{sentiment} identity is recognized on these three data sets.}
		\label{tab::tc-2}
		\begin{center}
			\scalebox{1.0}{
				\begin{tabular}{c|c|c|c|c|c|c}\hline
					& \multicolumn{2}{c|}{\textsc{Twitter}} & \multicolumn{2}{|c}{\textsc{MR}}& \multicolumn{2}{|c}{\textsc{Treebank}}\\\hline
					Algorithm&  Micro-F1& Macro-F1 &  Micro-F1& Macro-F1 &  Micro-F1& Macro-F1 \\ \hline \hline
					BOW& 	75.27	     &     	75.27	 &  71.90    &  71.90    &  \textbf{79.51}     &  \textbf{ 79.47}    \\ \hline				
					LDA& 65.33&65.32 &56.74&56.74 &63.28 &63.28\\ \hline
					Skipgram&73.00 &73.02 &67.05 &67.05 &72.76 & 72.76 \\ \hline 
					LINE&73.18 &73.19 &71.06 &71.06 &74.79 &74.78  \\ \hline 
					MSSG&72.98 &72.97 &65.00 &64.09 &71.39 &71.39\\ \hline 
					TWE(sentiment)&72.67 &72.67&70.81&70.82&74.95 &74.94  \\ \hline
					ISE(sentiment)&\textbf{75.50} &\textbf{75.50} &\textbf{73.52} &\textbf{73.51} &78.47 &78.43  \\  \hline 											
				\end{tabular}
			}
		\end{center}
	\end{table*}

Let us start by looking at the results of text classification. Table~\ref{tab::tc-1} reports the results of text classification on \textsc{20NG}, \textsc{DBLP} and \textsc{WikiSample} data sets. On these data sets, the \emph{topic} and \emph{category} identities are recognized. Let us first compare the results of the approaches without considering the word meanings in different contexts including Skipgram and LINE. We can see that the performance of LINE consistently outperforms Skipgram on all the three data sets, which is consistent with the results reported in~\cite{tang2015line,tang2015pte}. The reason is that the LINE model learns the word embedding through first constructing the word co-occurrence network and then embedding the network, which captures more global structures of word co-occurrences compared to free text. 

The LDA, MSSG, TWE, ISE models all considers the variations of word meanings in different contexts. In the LDA, the same word may be assigned to different topics in different documents; in the MSSG, each word can take different senses in different contexts; in TWE and ISE, the meanings of the words are determined by their identities, which are different depending on the contexts. 

We can see that though the LDA is able to differentiate the meanings of words in different contexts by inferring the topic identities, the performance is not so good. This is because the meanings of words assigned to the same topic are assumed to be totally the same, which may not be reasonable. Our approach effectively complements this through identity-sensitive word embedding, which differentiates the meanings of words in the same topics/identities. The performance of MSSG is even inferior to that of Skipgram. The reason may be that the MSSG dynamically infer multiple senses of the words and it may be very hard to accurately estimate the senses of the words in different contexts during the learning process. For both the TWE and our ISE models, when \emph{topic} identity is considered, the performance outperforms both the Skipgram and LINE.

Our ISE model also consistently outperforms the TWE model. The reasons are twofold. \emph{First,} our ISE approach leverages the identity-level word co-occurrences while the TWE does not utilize this information; \emph{second}, our ISE model learns the word embeddings through embedding the heterogeneous networks, which capture global structures of word co-occurrences compared to free text. We also notice that the performance of TWE becomes very bad when using the category identity, even inferior to that of Skipgram. The reason is that the category identities of the tokens in the test documents are very hard to recognize by the TWE model.  However, our ISE model can effectively recognize the identities of the tokens in the test document through the context embeddings of words, which are not available in the TWE model. We also notice that the results of ISE with \emph{category} identity are better than with \emph{topic} identity. The reason is that the \emph{category} identity is more relevant to the text classification task compared to the \emph{topic} identity.  Our ISE model also outperforms the BOW method, the dimension of which is much larger. 

Table~\ref{tab::tc-2} reports the results of text classification on the \textsc{Twitter}, \textsc{MR}, and \textsc{Treebank} data sets. On the three data sets, the \emph{sentiment} identity is recognized. Similar results can be observed compared to the \textsc{20NG}, \textsc{DBLP} and \textsc{WikiSample} data sets. The performance of LDA is still not good. The LINE model consistently outperforms the Skipgram. The performance of MSSG is still inferior to the Skipgram. The performance of TWE with sentiment identity is also not good, even inferior to the LINE model. Our ISE model performs the best by considering the sentiment identity (on \textsc{Treebank} dataset, a little bit inferior to BOW, which uses a much larger feature dimension).

	\begin{figure}[!htdb]
		\centering  
		\subfigure[\textsc{20NG}]{
			\includegraphics[width=0.22\textwidth]{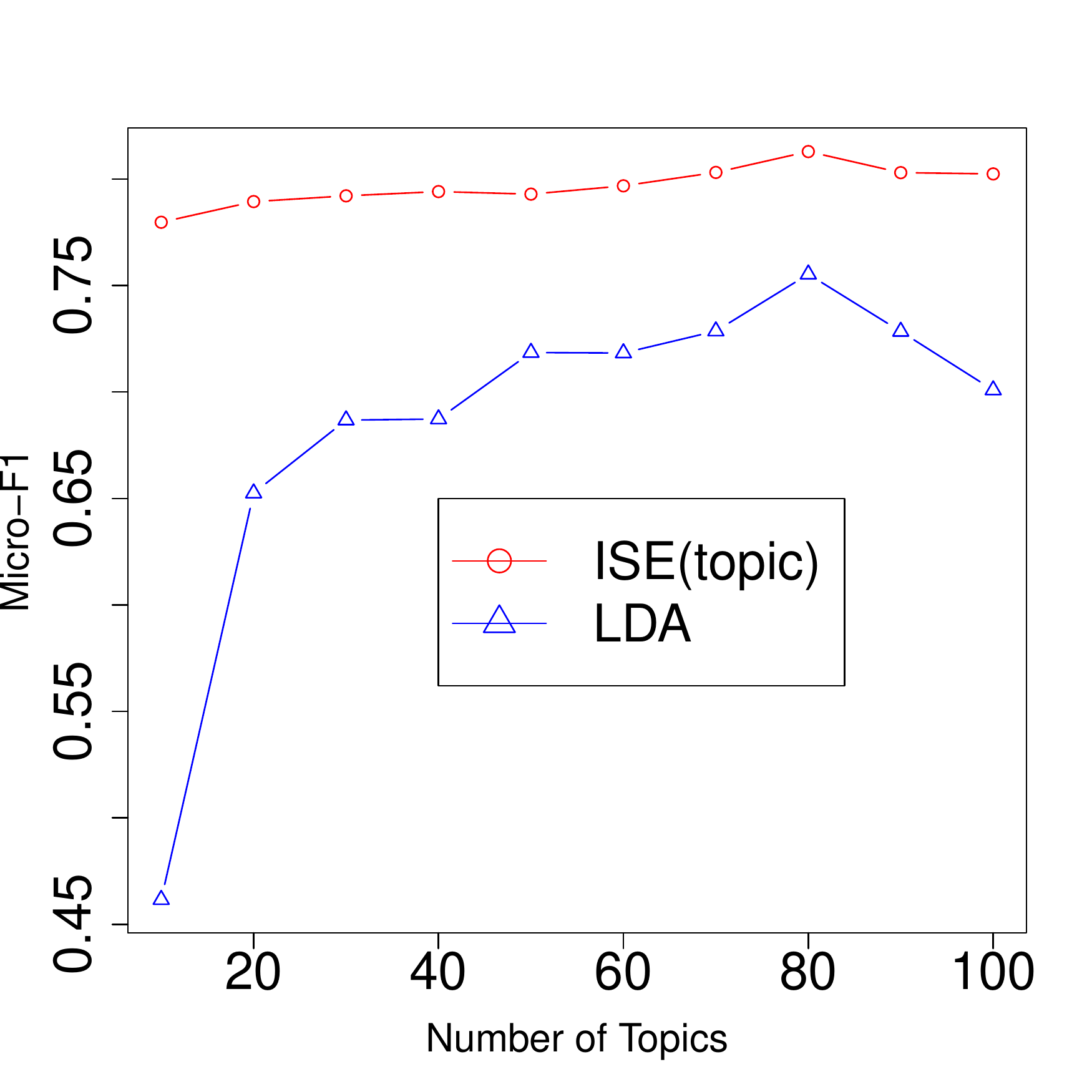}
		}
		\subfigure[\textsc{DBLP}]{
			\includegraphics[width=0.22\textwidth]{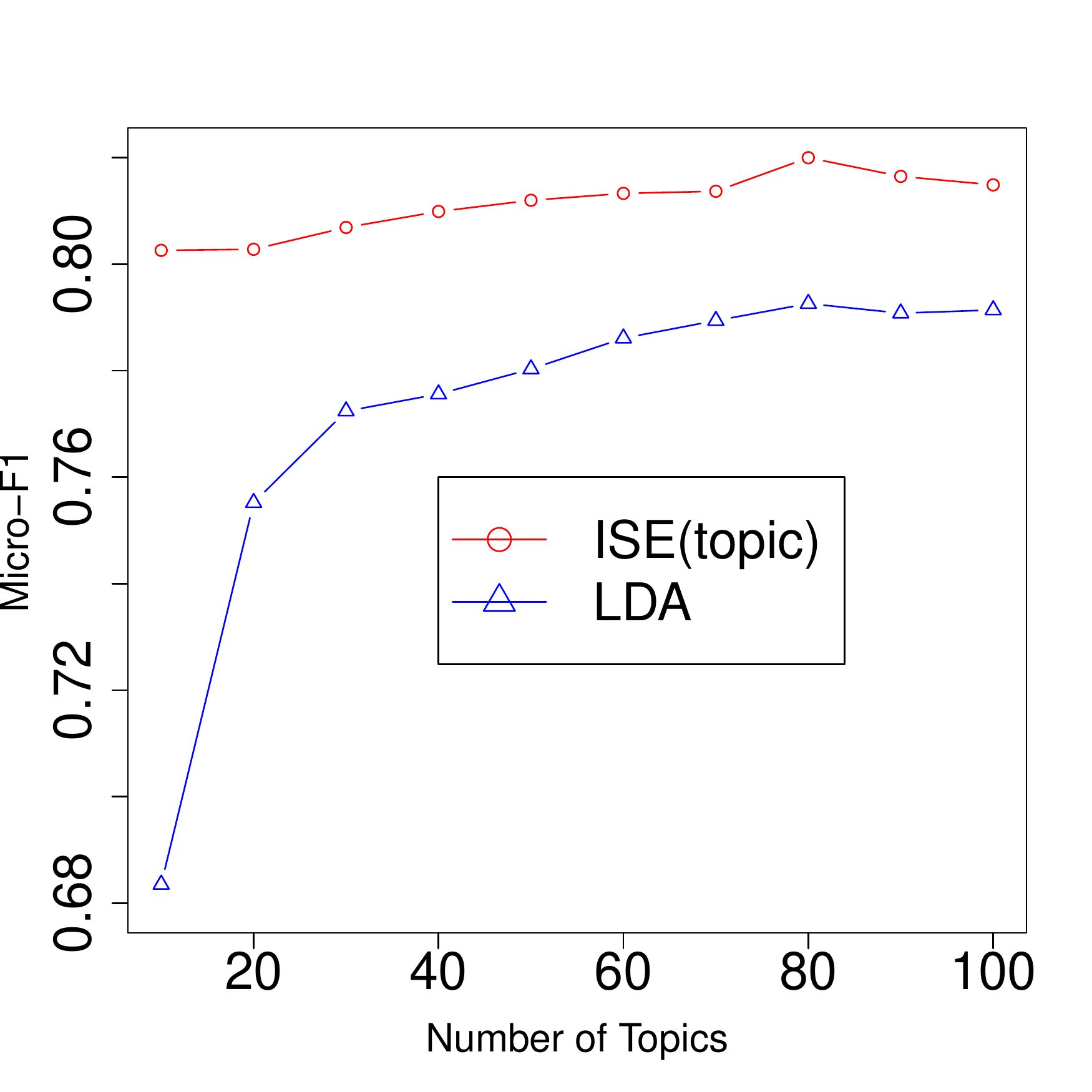}
		}
		\caption{Performance of ISE w.r.t the number of identities (topics). }
		\label{fig::sensitivity}		 				
	\end{figure}
	
We are also interested in how the number of identities would affect the performance. Fig.~\ref{fig::sensitivity} reports the performance of text classification w.r.t the number of identities (topics) on the 20NG and DBLP data sets. We can see that compared to the LDA, the performance of the ISE model is quite stable and not sensitive to the number of topics, which saves the overhead of choosing the appropriate number of topics when using the LDA model.  The performance gain over LDA is especially significantly when the number of topics specified is less than the optimum one. The reason is that the LDA does not differentiate the meanings of words in the same topic/identity, which is very serious when the number of topics is small, while in our ISE model the meanings of words with the same topic/identity are differentiated through embedding.

\subsubsection{Contextual Word Similarity}

\begin{table}[!htdb]
	\caption{Results of contextual word similarity by training word embeddings on the \textsc{WikiFull} data set. \emph{Topic} identity is utilized.}
	\label{tab::results-cws}
	\begin{center}
		\scalebox{1.0}{
			\begin{tabular}{c|c|c}  \hline
				Algorithm& \multicolumn{2}{c}{Spearman Correlation} \\ \hline \hline
				& d=50 & d=300  \\ \hline
				MSSG& 49.17& 57.26 \\ \hline
				TWE(topic)& 57.60 &61.23\\ \hline
				ISE(topic)& \textbf{58.54} &\textbf{62.27}\\ \hline
			\end{tabular}	
		}
	\end{center}
\end{table}

Next, let us look at the results on the task of contextual word similarity. Table~\ref{tab::results-cws} reports the results on the \textsc{WikiFull} data set. We only compare the word embedding approaches that consider contextual word meanings including MSSG, TWE and ISE. For the TWE and ISE models, the \emph{topic} identity is used as it can be estimated in an unsupervised way through LDA and 200 topics are estimated. We can see that the performance of TWE and ISE significantly outperforms the MSSG model, which is consistent with previous results on text classification. The ISE still outperforms TWE as it leverages identity-level word co-occurrences and represents different levels of word co-occurrences through heterogeneous networks. 

\subsection{Nearest Neighbors}

To intuitively compare the embeddings learned by different approaches, in this part we look at the nearest neighbors of words based on the embeddings. The similarity of the words are calculated as the cosine similarity of the word embeddings. Table~\ref{tab::Wiki-cs} presents some examples of nearest neighbors based on the embeddings learned by different approaches on the \textsc{WikiFull} data set.  Overall, we can see that the word embeddings learned by our ISE approach are indeed able to capture different meanings of the same word. Take the word ``language'' as an example. The three different meanings of ``language'' recognized by our ISE model are ``linguistic'', ``programming language'' and  ``human language''. The MSSG model only discovers two different meanings. The result of Skipgram is even worse, which is a mixture combination of multiple meanings.

 \begin{table*}
 	\caption{Comparison of nearest neighbors based on embeddings learned by different approaches on the \textsc{WikiFull} data set. }
 	\label{tab::Wiki-cs}
 	\centering
 	\scalebox{0.7}{
 		\begin{tabular}{c|c|c|c|c|c} \hline
 			Word& Algorithm  & Most similar words &Word& Algorithm  & Most similar words \\ \hline \hline
 			\multirow{6}{0.6in}{language}&	\multirow{3}{0.6in}{ISE(topic)}& multilingual, speaking, vocabulary, write, linguistic &\multirow{6}{0.6in}{bank}&	\multirow{3}{0.6in}{ISE(topic)}& branch, trust, commercial, savings, deposit\\ 
 			&	&programming, java, compiler, implementation, oriented &	& &river, opposite, mouth, crossing, bridge\\ 
 			&	&Hindi, Malayalam, Bengali, Marathi, Oriya  &	& &\\ \cline{2-3} \cline{5-6}
 			& \multirow{2}{0.6in}{MSSG}  &programming, procedural, lisp, java, object & &\multirow{2}{0.6in}{MSSG} &repurchases, securitizes, money, credit, stock  \\
 			&   &vernacular, English, speakers, siswati, berber  &	& &thouet, swale, north, east, side\\ \cline{2-3} \cline{5-6}
 			& \multirow{1}{0.6in}{Skipgram}  & multilingual, englishmandarin, spoken, linguistic, english & &\multirow{1}{0.6in}{Skipgram} & citibank, HSBC, vanquish, vanquish, postbank \\ \hline    
			\multirow{5}{0.6in}{star}&	\multirow{2}{0.6in}{ISE(topic)}& sky, phoenix, globe, golden, blue &\multirow{5}{0.6in}{windows}&	\multirow{2}{0.6in}{ISE(topic)}& server, operating, version, Vista, XP\\ 
			&	&series, movies, comics, prequel, film  &	& &rear, floor, large, roof, entrance\\  \cline{2-3} \cline{5-6}
			& \multirow{2}{0.6in}{MSSG}  &nebula, orion, giant, moon, planet & &\multirow{2}{0.6in}{MSSG} & vista, os, XP, MS, dos \\
			&   &episode, batman, prequel, series, movie  &	& &doors, openings, arcade, freestanding, sockets\\ \cline{2-3} \cline{5-6}
			& \multirow{1}{0.6in}{Skipgram}  &sunlike, gazers, hypergiant, supergiant, galaxy  & &\multirow{1}{0.6in}{Skipgram} &microsoft, desktop, console, linux, cygwin \\ \hline 	
			\multirow{5}{0.6in}{left}&	\multirow{2}{0.6in}{ISE(topic)}& leaving, moving, position, coming, return &\multirow{5}{0.6in}{fox}&	\multirow{2}{0.6in}{ISE(topic)}&  nba, abc, cbs, espn, news\\ 
			&	&upper, middle, facing, side, top &	& &hawk, eagle, buck, bird, duck\\  \cline{2-3} \cline{5-6}
			& \multirow{2}{0.6in}{MSSG}  &leaving, returned, stayed, joined, leave & &\multirow{2}{0.6in}{MSSG} & wolf, bear, ringneck, bullfrogs, antelope \\
			&   &right, radishes, centerline, backcourts, kingside  &	& &upn, fxm, nbc, wfft, kmov\\ \cline{2-3} \cline{5-6}
			& \multirow{1}{0.6in}{Skipgram}  &right, leaving, coming, arrive, leave  & &\multirow{1}{0.6in}{Skipgram} &cbs, nbc, abc, ktvi, wtxf \\ \hline 						
			\multirow{7}{0.6in}{apple}&	\multirow{3}{0.6in}{ISE(topic)}& strawberry, apples, plum, blueberry, cherry &\multirow{7}{0.6in}{network}&	\multirow{3}{0.6in}{ISE(topic)}&  areas, exchange, distributed, services, link\\ 
			&	&iphone, ipod, ipad, store, ios  &	& &abc, programming, broadcast, stations, nationwide\\ 
			&	&macintosh, mac, computers, server, products  &	& &rail, lines, stations, traffic, routes\\  \cline{2-3} \cline{5-6}			
			& \multirow{3}{0.6in}{MSSG}  &pc, macintosh, IBM, Microsoft, Intel & &\multirow{3}{0.6in}{MSSG} & connection, wan, routing, noncentralized, protocol, ip \\
			&   &pear, honey, pumpkin, potato, nut   &	& &channel, cable, tv, broadcaster, newsasia\\
			&   &Microsoft, activision, sony, retail, gamestop    &	& &\\ \cline{2-3} \cline{5-6}			
			& \multirow{1}{0.6in}{Skipgram}  &blackberry, macintosh, appleworks, ibook, iwork  & &\multirow{1}{0.6in}{Skipgram} &citynet, cablecast, cable, csnet, channel \\ \hline 												
 		\end{tabular}
 	}
 \end{table*}

\subsection{Visualization of Word Embeddings}
	\begin{figure*}[!htdb]
		\centering     
		\subfigure[Overview of the identity-sensitive word embeddings\label{fig::visualization-overview}]{
			\includegraphics[width=1\textwidth]{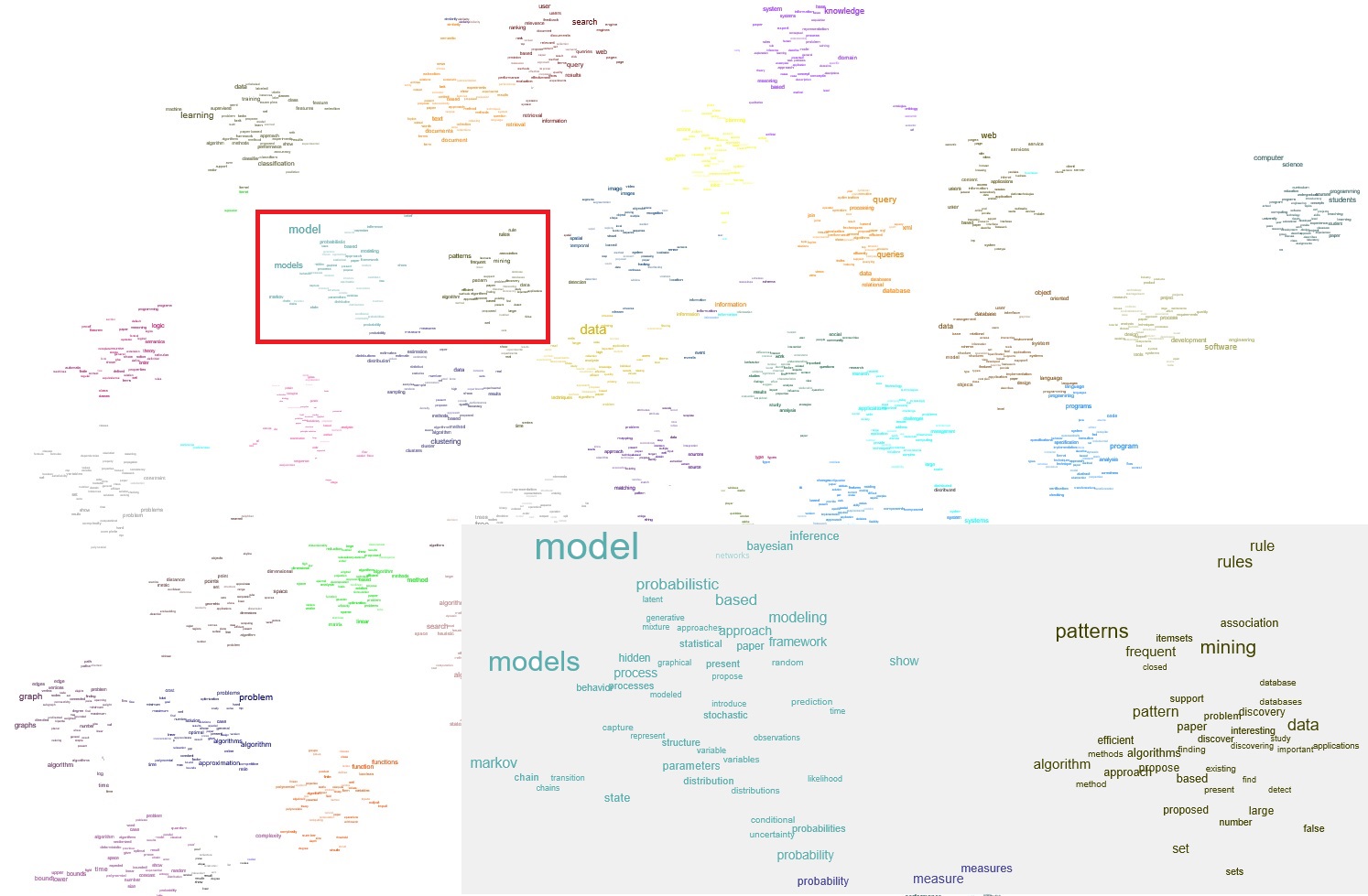}
		}
		\subfigure[The meanings of word ``networks''\label{fig::visualization-networks}]{
			\includegraphics[width=0.4\textwidth]{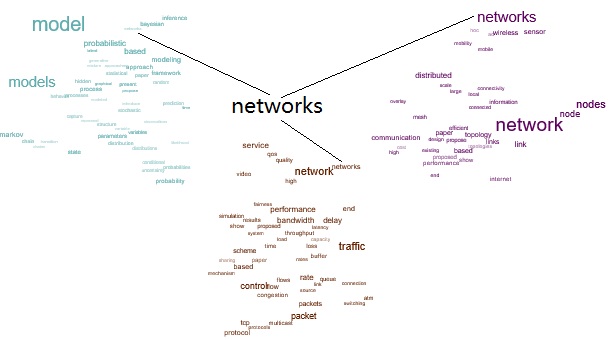}
		}
		\subfigure[The meanings of word ``learning''\label{fig::visualization-learning}]{
			\includegraphics[width=0.4\textwidth]{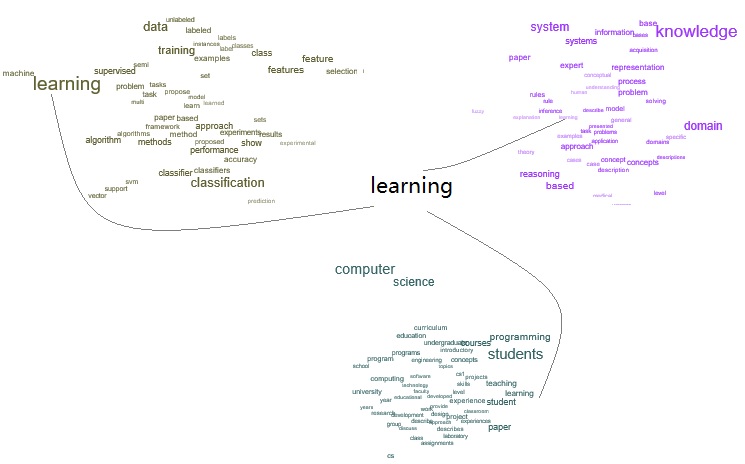}
		}	
		\subfigure[The meanings of word ``language''			\label{fig::visualization-language}]{
			\includegraphics[width=0.4\textwidth]{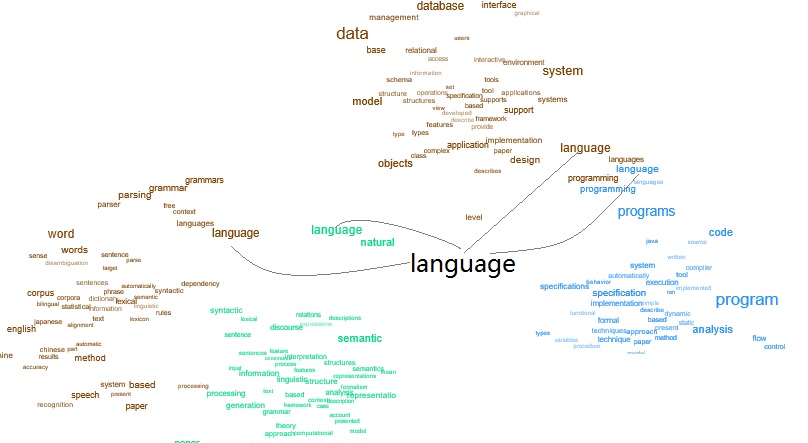}
		}
		\subfigure[The meanings of word ``semantic''			\label{fig::visualization-semantic}]{
			\includegraphics[width=0.4\textwidth]{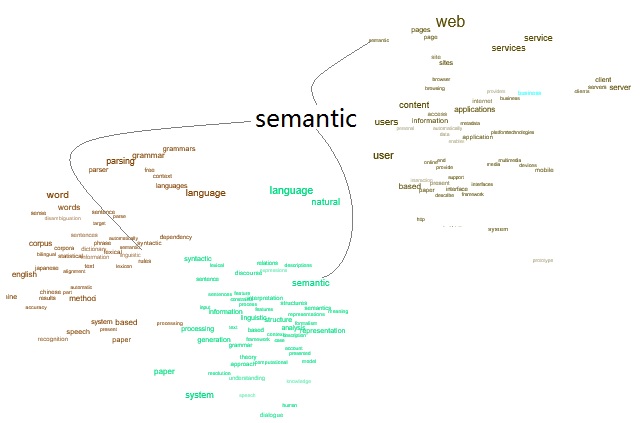}
		}						

		\caption{Visualization of the identity-sensitive word embeddings on the DBLP data set using the tSNE~\cite{van2008visualizing}. \emph{Topic} identity is used. Different colors represent different identities 
			(topics). }
		\label{fig::vis-dblp}
	\end{figure*}

Fig.~\ref{fig::vis-dblp} shows a visualization of our identity-sensitive word embeddings on the \textsc{DBLP} data set with topic identity. The words are mapped to 2-dimensional space with the tSNE~\cite{van2008visualizing}. Fig.~\ref{fig::visualization-overview} gives an overview of the visualization. Different colors represent different topics/identities, and the top 50 ranked words in each topic are visualized. At the right bottom, we zoom in a part of the visualization and present two topics. Overall, we can see that the identity-sensitive word embeddings have a very clear semantic structure. Words belonging to the same identities/topics are clustered together and different identities/topics are well separated. At the bottom of the two topic examples in Fig.~\ref{fig::visualization-overview}, we find that the word ``probability'' taking on different identities/topics are very close to each other. This may be explained by the linguistic theory that the meanings of words with different senses are still very close to each other~\cite{chomsky1988current}.

Fig.~\ref{fig::visualization-networks}, ~\ref{fig::visualization-learning}, ~\ref{fig::visualization-language}, ~\ref{fig::visualization-semantic} show some examples of words that appear in multiple topics (or have multiple identities) including the word ``networks'', ``learning'', ``language'', and ``semantic''. For example, the word ``networks'' can refer to ``Bayesian networks'' or ``wireless networks''; the word ``language'' can refer to ``programming language'' or ``natural language''; the word ``learning'' can refer to ``machine learning'' or `` human learning''.

%% file: conclusion.tex
\section{Conclusion}
\label{sec::conclusion}
In this paper, we studied the problem of contextual word embeddings. We acknowledged that the same words can take on different identities in different contexts, and proposed to learn identity-sensitive word embeddings. We proposed to represent different levels of word co-occurrences with a heterogeneous network. A principled heterogeneous network embedding approach based on our previous work is further proposed, which is able to obtain identity-sensitive word embeddings and the embeddings of word identities. Experiments show that the identity-sensitive word embeddings learned by our approach are indeed able to capture different meanings of the same word. Experiments on text classification and contextual word similarity proved the effectiveness of our proposed approach.

In the future, we plan to study more types of word identities such as POS tags and ideologies. Besides, an interesting problem to investigate is that whether the identity-sensitive word embeddings can be useful in improving the performance of recognizing the identities of the word tokens in different contexts. For example, whether the identity-sensitive word embeddings can be useful for topic modeling process.  